\documentclass[final]{cvpr}

\usepackage{times}
\usepackage{epsfig}
\usepackage{graphicx}
\usepackage{amsmath}
\usepackage{amssymb}
\usepackage{tabularx}
\usepackage[T1]{fontenc}
\usepackage{arydshln}
\usepackage[ruled,vlined,linesnumbered]{algorithm2e}
\usepackage[pagebackref=true,breaklinks=true,colorlinks,bookmarks=false]{hyperref}

\SetCommentSty{mycommfont}

\def\cvprPaperID{6014} % *** Enter the CVPR Paper ID here
\def\confYear{CVPR 2021}

\begin{document}

%%%%%%%%% TITLE
\title{From Synthetic to Real: Unsupervised Domain Adaptation \\ for Animal Pose Estimation}

\author{Chen Li \quad \quad Gim Hee Lee\\
Department of Computer Science, National University of Singapore\\
{\tt\small \{lic, gimhee.lee\}@comp.nus.edu.sg}
}

\maketitle

%%%%%%%%% ABSTRACT
\begin{abstract}
   Animal pose estimation is an important field that has received increasing attention in the recent years. The main challenge for this task is the lack of labeled data. 
   Existing works circumvent this problem with pseudo labels generated from data of other easily accessible domains such as synthetic data. However, these pseudo labels are noisy even with consistency check or confidence-based filtering due to the domain shift in the data. To solve this problem, we design a multi-scale domain adaptation module (MDAM) to reduce the domain gap between the synthetic and real data. We further introduce an online coarse-to-fine pseudo label updating strategy. Specifically, we propose a self-distillation module in an inner coarse-update loop and a mean-teacher in an outer fine-update loop to generate new pseudo labels that gradually replace the old ones.    
   Consequently, our model is able to learn from the old pseudo labels at the early stage, and gradually switch to the new pseudo labels to prevent overfitting in the later stage. We evaluate our approach on the TigDog and VisDA 2019 datasets, where we outperform existing approaches by a large margin. We also demonstrate the generalization ability of our model by testing extensively on both unseen domains and unseen animal categories.  Our code is available at the project website\footnote{\url{https://github.com/chaneyddtt/UDA-Animal-Pose}}.
\end{abstract}

%%%%%%%%% BODY TEXT
\section{Introduction}

\begin{figure}
    \centering
    \includegraphics[width=0.96\linewidth]{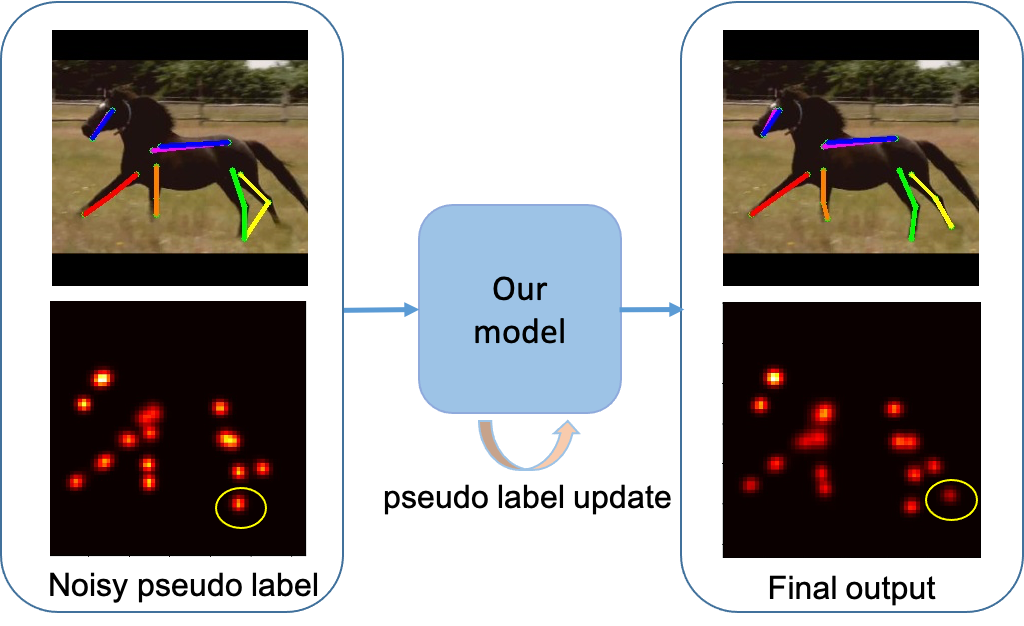}
    \caption{Our method takes in noisy pseudo labels (\eg hind hoof on left image) generated from model trained with labeled synthetic data and outputs the correct animal pose on real images.
    %Top: Some examples of wrong predictions, the left front  hoof (left), the right back hoof (middle) and the left back hoof (right). Bottom: The confidence score for each joint prediction, where the brighter dot denotes more confident prediction. Note that we remove the confidence score for a correct prediction if it is near to a wrong prediction for clarity, \eg right front hoof in the first heatmap.
    }
    \vspace{-3mm}
    \label{fig:teaser}
\end{figure}

Animal pose estimation has received increasing attention over the last few years because of many potential applications in zoology, biology and aquaculture. Despite the great success of applying deep neural networks to human pose estimation, the lack of well-labeled animal pose data makes it infeasible to directly leverage on the powerful deep learning approaches. Existing 
works overcome this problem by transferring knowledge from other more accessible domains such as synthetic animal data \cite{mu2020learning, biggs2018creatures, zuffi2019three,zuffi2018lions, Zuffi_2017_CVPR} or human data \cite{cao2019cross}. The advantage of synthetic data is that it is low cost and convenient to generate a large scale of data with accurate ground truth. Moreover, the domain gap between synthetic and real animals is more manageable than that between other domains such as human and animals. This is evident from the results of \cite{cao2019cross}, where sufficient labeled data in the real animal domain is needed for the network to work despite the use of sophisticated domain adaptation techniques. %where labeled data in the real animal domain is needed for the network to show decent performance since training on human pose alone totally fails despite the use of sophisticated domain adaptation techniques. 

The domain gap between synthetic and real animals mainly comes from the differences in texture and background, and the limited pose variance of synthetic data. To solve the domain shift problem, existing works first generate pseudo labels with a model trained on synthetic data, and then gradually incorporate more pseudo labels into the training according to the confidence score. However, these pseudo labels are inaccurate even with refinement techniques such as confidence-based filtering \cite{cao2019cross} or geometry-based consistency check \cite{mu2020learning}. 
% Fig.~\ref{fig:teaser}
% shows that the model trained on synthetic animals gives wrong predictions with high confidence, and cannot be alleviated by the geometry-based consistency check. These noisy pseudo labels lead to degraded performance when used naively for training.
Fig.~\ref{fig:teaser} shows an example where a model trained on synthetic animals gives wrong predictions (\eg the hind hoof) with high confidence (marked in yellow circle in the heatmap). This kind of noisy pseudo labels cannot be filtered out based on the confidence score and will lead to degraded performance when used naively for training.

In this paper, we propose a novel approach to learn from synthetic animal data. We design a multi-scale domain adaptation module (MDAM) to reduce the domain gap. Our MDAM consists of a pose estimation module and a domain classifier. %The pose estimation module takes the image as input, and output a heatmap for the animal pose. We concatenate the feature maps at multiple scales and feed it into the domain discriminator to close the domain gap. 
We first train the pose estimation module with the synthetic data \cite{mu2020learning} to generate an initial set of pseudo labels for the real animal images. We then train our MDAM on the synthetic labels and the pseudo labels. However, the accuracy of MDAM is limited by the presence of noise in the pseudo labels. 
To alleviate this problem, we introduce an online coarse-to-fine pseudo label updating strategy. Specifically, we propose a self-distillation module in the inner coarse-update loop and a mean-teacher \cite{tarvainen2017mean} in the outer fine-update loop to generate better pseudo labels that gradually replace the old noisy ones. 

We design our pseudo label updating strategy according to the \textit{memorization effect} \cite{pmlr-v70-arpit17a, ZhangBHRV17} of deep networks, which states that deep networks learn from clean samples at the early stage before eventually memorizing (\ie overfits to) the noisy ones. To avoid the memorization effect, we rely more on the initial pseudo labels at the early stage when the self-distillation module and mean-teacher are still at their infancy in training. Our coarse-to-fine pseudo label updating strategy gradually replaces the noisy initial labels when the self-distillation module and mean-teacher gained
enough competency to generate more reliable pseudo labels. Consequently, we are able to supervise our network with more accurate pseudo labels and prevent overfitting at the same time. As illustrated in Fig.~\ref{fig:teaser}, our model can successfully locate the joint (hind hoof on the right image) although the initial pseudo label is not accurate.

We validate our approach on the TigDog Dataset \cite{del2015articulated}, where we outperform existing unsupervised domain adaptation techniques by a large margin. We also demonstrate the generalization capacity of our approach by directly testing on the Visual Domain Adaptation Challenge dataset (VisDA2019), the Zebra dataset \cite{zuffi2019three} and the Animal-Pose dataset \cite{cao2019cross}. Experimental results show that our approach can generalize well to both unseen domains and unseen animal categories. Our main contributions are as follows:
\begin{itemize}
    \item We design an unsupervised domain adaptation pipeline for animal pose estimation, which consists of a multi-scale domain adaptation module, a self-distillation module and a mean-teacher network. 
    \item We propose an online coarse-to-fine pseudo label updating strategy to alleviate the negative effect of unreliable pseudo labels.
    \item Our approach achieves state-of-the-art results on the TigDog dataset and the VisDA2019 dataset, and can also generalize well to unseen domains and unseen animal categories.
\end{itemize}

\section{Related Work}
\paragraph{Human Pose estimation.} Human pose estimation has been an active research field for decades. One of the most popular early approaches is the pictorial structure \cite{dantone2013human, andriluka2009pictorial, sapp2010adaptive} which uses a tree structure to model the spatial relationships among body parts. These methods do not perform well in complex scenarios because of the limited representation capabilities. Recently, deep learning based approaches \cite{pishchulin2016deepcut, newell2016stacked, chu2017multi, yang2017learning, wei2016convolutional, chen2018cascaded, xiao2018simple, papandreou2017towards} have achieved significant progress due to the availability of large scale training data such as the MPII dataset \cite{andriluka20142d} and the COCO keypoint detection dataset \cite{lin2014microsoft}.  Existing works can be divided into two categories. The first category \cite{chen2018cascaded, xiao2018simple, papandreou2017towards} adopts a single stage backbone network, typically ResNet \cite{he2016deep}, to generate deep features, after which upsampling or deconvolution is applied to generate heatmaps with higher spatial resolution. The second category \cite{newell2016stacked, chu2017multi,yang2017learning, wei2016convolutional} is based on a multi-stage architecture where the generated results from the previous stage are refined step by step. In this paper, we adopt the single stage approach as our basic structure so that we can directly apply domain adaptation to the output of the backbone network.

\vspace{-3mm}
\paragraph{Animal Pose Estimation.}
Animal pose estimation is relatively under-explored compared to human pose estimation mainly due to the lack of labeled data. To solve this problem, Mu \etal \cite{mu2020learning} use synthetic animal data generated from CAD models to train their model, which is then used to generate pseudo labels for the unlabeled real animal images. Subsequently, the generated pseudo labels are gradually incorporated into training based on three consistency check criteria. Cao \etal \cite{cao2019cross} propose a cross-domain adaptation scheme to learn a shared feature space between human and animal images such that their network can learn from existing human pose datasets. They also select pseudo labels into the training based on the confidence score. In contrast to \cite{mu2020learning} which does not need any labels for real animal images, \cite{cao2019cross} needs 
% 2000 (or 4000) out of a total of 5117 images 
part of the real animal images  to be labeled in their dataset to facilitate a successful transfer. Similar to \cite{mu2020learning}, we focus on unsupervised domain adaptation from synthetic animal data. Instead of gradually incorporating pseudo labels into training, we conduct an online coarse-to-fine pseudo label update to alleviate the negative effect of noisy pseudo labels. 

In addition, there are also several works focusing on 3D animal pose and shape estimation \cite{Zuffi_2017_CVPR, zuffi2018lions, zuffi2019three, biggs2018creatures, kanazawa2016learning, biggs2020wldo, }. \cite{Zuffi_2017_CVPR} builds a statistical 3D shape model SMAL by learning from scans of toy animals. To recover more detailed 3D shape of animals, \cite{zuffi2018lions} regularizes the deformation of the mesh from SMAL to constrain the final shape. \cite{zuffi2019three} trains a neural network on a digitally generated dataset to predict 3D pose, shape and texture for the SMAL model. 

\vspace{-3mm}
\paragraph{Unsupervised Domain Adaptation.}
Unsupervised domain adaptation aims to learn a model from a labeled source domain which can perform well on an unlabeled target domain. One mainstream approach is based on adversarial learning \cite{ganin2015unsupervised, hoffman2018cycada, volpi2018adversarial, wu2020dual}, where a feature extractor tries to learn domain-invariant features in order to fool a domain discriminator. The alignment with adversarial learning can facilitate the transfer of labels from the source to the target domain. Besides feature level alignment, other works also try to reduce the domain shift in the input \cite{hoffman2018cycada} or output level \cite{tsai2018learning, yang20183d}. In this paper, we apply a domain classifier to the feature maps of multiple scales such that both global and local features can be aligned. 

\vspace{-3mm}
\paragraph{Learning from Noisy Data}
Learning from noisy labels is an important research topic especially for the deep learning community. This is because deep learning algorithms rely heavily on large scale labeled training data that is costly to collect. %while accurate annotations are expensive to collect. 
To reduce the negative effect of noisy labels, some approaches focus on training noise robust models by designing robust losses \cite{ghosh2017robust,wang2019symmetric,zhang2018generalized} or by correcting the loss with a transition matrix \cite{patrini2017making, goldberger2016training, xia2019anchor}. Sample selection based approaches \cite{malach2017decoupling, jiang2018mentornet, han2018co, yu2019does} attempt to select possibly clean samples in each iteration for training. One of the most representative methods is Co-teaching \cite{han2018co, yu2019does}, which trains on all samples at the beginning and gradually drops the samples with large loss values. This \textit{small-loss} trick , which is based on the \textit{memorization effect} \cite{pmlr-v70-arpit17a, ZhangBHRV17} of deep networks, has also adopted by other works \cite{jiang2018mentornet, song2019selfie} to select more reliable labels. Given the noisy pseudo labels, we also conduct sample selection similar to the Co-teaching. Moreover, We gradually update the pseudo labels with the knowledge from a self-distillation module and a teacher network.

\section{Our Method}
\begin{figure*}[t!]
    \centering
    \includegraphics[width=0.98\textwidth]{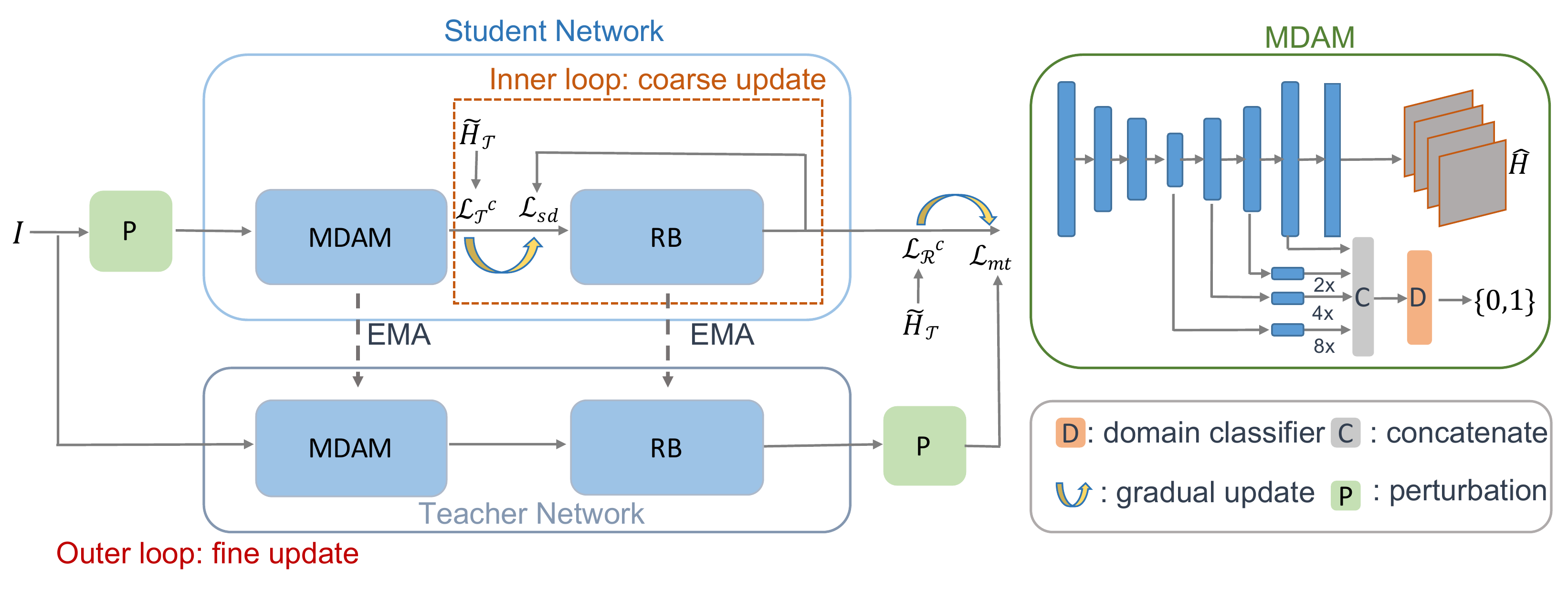}
    \vspace{-2mm}
    \caption{Our network is a student-teacher architecture, where the student network consists of a multi-scale domain adaptation module (MDAM), a refinement block (RB) and a self-feedback loop. We conduct online coarse-to-fine pseudo label update through the inner loop and the outer loop, respectively.}
    \label{fig:network}
\end{figure*}

We propose an unsupervised domain adaptation approach for animal pose estimation. The labeled source domain $\mathcal{S}$ consists of synthetic animal images $I_{\mathcal{S}}$ and the corresponding pose labels $Y_{\mathcal{S}}$ generated from CAD models, and the unlabeled target domain $\mathcal{T}$ consists of in-the-wild animal images $I_{\mathcal{T}}$ without pose labels. The goal is to learn a pose estimation model that can adapt well to the unlabeled target domain. To this end, we design a student-teacher network as shown in Fig.~\ref{fig:network}. The student and teacher networks share the same architecture: a basic pose estimation module (PEM), a self-distillation module (SDM) and a domain classifier (DC). We first pretrain the PEM on $I_\mathcal{S}$ and use it to generate pseudo labels for $I_{\mathcal{T}}$. However, those pseudo labels are noisy due to the domain gap between the synthetic and real images, and can hurt the performance when used naively in training. To alleviate this negative effect, we propose an online coarse-to-fine pseudo label updating strategy with the self-distillation module and teacher network.
%incorporating knowledge distillation from the RM and teacher network. 

\subsection{Multi-scale Domain Adaptation Module}
Our MDAM consists of a pose estimation module and a domain classifier $D$. The pose estimation module follows an encoder-decoder architecture, where the encoder is the feature extractor $G$ and the decoder is the pose estimator $P$. Given a pair of images $(I_\mathcal{S},I_\mathcal{T}) \in \mathbb{R}^{H \times W \times 3}$ from the source and target domains, we feed them into the pose estimation module to get the corresponding feature maps $(F_\mathcal{S}, F_\mathcal{T})$ and heatmaps $(\hat{H}_\mathcal{S}, \hat{H}_\mathcal{T})$:
\begin{equation}
\begin{split}
    \label{ftmaps}
    & F_\mathcal{S} = G(I_\mathcal{S}), \quad \hat{H}_\mathcal{S} = P(F_\mathcal{S}), \\
    & F_\mathcal{T} = G(I_\mathcal{T}), \quad \hat{H}_\mathcal{T} = P(F_\mathcal{T}).
\end{split}
\end{equation}

Similar to human pose estimation \cite{newell2016stacked}, we define the animal pose estimation loss in the source domain as the mean-square error (MSE) between the estimated and ground truth heatmaps:
\begin{align}
\label{eq: pose_source}
    \mathcal{L}_\mathcal{S} = \frac{1}{\mathcal{N}} \sum_{i,j,c}\|\hat{H}_\mathcal{S} (i,j,c) - H_\mathcal{S} (i,j,c)\|^2,
\end{align}
where $\mathcal{N}=h_o \times w_o \times K$, $H_\mathcal{S}$ represents the ground truth heatmaps with resolution $h_o \times w_o$ and $K$ represents the total number of joints.

We use the pseudo labels $\tilde{H}_\mathcal{T}$ for the target domain since the ground truth for the target domain is not available:
\begin{align}
\label{eq: pose_target}
    \mathcal{L}_\mathcal{T} = \frac{1}{\mathcal{N}} \sum_{i,j,c}\|\hat{H}_\mathcal{T} (i,j,c) - \tilde{H}_\mathcal{T} (i,j,c)\|^2.
\end{align}
Note that these pseudo labels $\Tilde{H}_\mathcal{T}$ and their corresponding confidence scores $C_\mathcal{T}$ are generated from our pose estimation module pretrained on the source domain data following the training procedure from \cite{mu2020learning}.

%Note that we first pretrain the pose estimation module on the source domain with \eqref{eq: pose_source} and generate pseudo labels $\Tilde{H}_\mathcal{T}$ and the corresponding confidence score $C_\mathcal{T}$ in the same way as \cite{mu2020learning}.
% The losses for both source and target domain are jointly optimized when training the DAM:
% \begin{align}
%         \mathcal{L}_{pose} = \mathcal{L}_\mathcal{S} + \lambda_{\mathcal{T}}\mathcal{L}_\mathcal{T},
% \end{align}
% where $\lambda_{\mathcal{T}}$ is the weight for the target domain.

To bridge the domain gap between the source and target domains, we apply a domain classifer $D$ \cite{ganin2015unsupervised, hoffman2018cycada, volpi2018adversarial} to the output of the feature extractor $G$. The domain classifier attempts to classify the real target data from the synthetic source data using a cross-entropy loss $\mathcal{L}_d$:
\begin{equation}
    \mathcal{L}_d = -\log (1-D(F_\mathcal{T})) - \log (D(F_\mathcal{S})),
    \label{eq:lossDomainClass}
\end{equation}
while the feature extractor tries to fool the domain classifier by maximize $\mathcal{L}_d$:
\begin{equation}
    \mathcal{L}_\text{adv} = -\mathcal{L}_d.
    \label{eq:lossDomainDisc}
\end{equation}
We use a gradient reversal layer \cite{ganin2015unsupervised} for optimization.

We apply the domain classifier to the feature maps at multiple scales 
given that both local (\eg a small batch around a joint) and global information (\eg the relationship between different joints) are important for joint detection. More specifically, we concatenate the intermediate outputs of the pose estimator and feed them into the domain classifier, as shown in the right part of Fig.~\ref{fig:network}.

\subsection{Coarse-to-Fine Pseudo Label Update}
The pseudo labels we use in Eq.~\eqref{eq: pose_target} are noisy although we filter the samples based on the consistency-check criteria described in \cite{mu2020learning}. To circumvent this problem, we propose the coarse-to-fine pseudo label updating strategy to gradually replace the noisy pseudo labels with more accurate ones. As shown Fig.~\ref{fig:network}, our coarse-to-fine pseudo label updating strategy consists of two nested loops. 

\vspace{-3mm}
\paragraph{Inner coarse-update loop:} As shown in Fig.~\ref{fig:network}, the inner loop consists of the self-distillation module: a refinement block(RB) and a self-feedback loop. The input to the refinement block is the output of MDAM $\hat{H}_\mathcal{T}$, and we denote its output as $\mathcal{R}_\mathcal{T}$. The output of MDAM is supervised by the output of the refinement block via the self-feedback loop with a self-distillation loss: 
\begin{equation}
    \mathcal{L}_\text{sd} = \frac{1}{\mathcal{N}} \sum_{i,j,c}\|\hat{H}_\mathcal{T}(i,j,c) -\mathcal{R}_\mathcal{T}(i,j,c)\|^2.
    \label{eq:lossSelfDist}
\end{equation}
We also supervise the output of MDAM $\hat{H}_\mathcal{T}$ concurrently with the noisy pseudo labels $\tilde{H}_\mathcal{T}$, \ie  
\begin{equation}
    \begin{split}
    & \mathbf{L}_\mathcal{T} = \frac{1}{|\mathcal{C}|} \sum_{c \in \mathcal{C}}\mathcal{L}^c_\mathcal{T}, \quad \text{where}~~\mathcal{C} = \{c \mid  \mathcal{L}_\mathcal{T}^{c} < l_\text{th}\},\\ 
    & \mathcal{L}^c_\mathcal{T} = \frac{1}{\mathcal{M}} \sum_{i,j}\|\hat{H}_\mathcal{T} (i,j,c) - \Tilde{H}_\mathcal{T} (i,j,c)\|^2.
    \end{split}\label{eq:lossPseudoMDAM}
\end{equation}
$\mathcal{M} = h_0 \times w_0$, and in contrast to Eq.~\ref{eq: pose_target}, $\mathcal{L}^c_\mathcal{T} \in \mathbb{R}^k$ do not sum over $c$, \ie $\mathcal{L}^c_\mathcal{T}$ is the loss term per joint. 
$\mathcal{C}$ is the set of joint indices with a loss value $\mathcal{L}^c_\mathcal{T}$ smaller than the threshold $l_\text{th}$, which dynamically decreases as the training proceeds. This means that we start the training with all the pseudo labels and gradually drop those with large loss values. The intuition is that the clean samples tend to exhibit smaller losses than noisy ones before the network eventually overfit to the noisy ones \cite{pmlr-v70-arpit17a, ZhangBHRV17}.
On the other hand, we assign a gradually increasing weight to $\mathcal{L}_\text{sd}$ in the total loss. This results in a net effect of gradually replacing the initial noisy pseudo labels with better pseudo labels produced by the refinement block $\mathcal{R}_\mathcal{T}$ at the later stage of training.  

\vspace{-3mm}
\paragraph{Outer fine-update loop:} As shown in Fig.~\ref{fig:network}, the outer loop is a student-teacher architecture. The student network consists of the multi-scale domain adaptation module and the self-distillation module. The teacher network has an identical architecture with the student network with the exception of the self-feedback loop in the self-distillation module. Furthermore, we follow the mean-teacher~\cite{tarvainen2017mean} paradigm to update the teacher model $\theta'$ with the exponential moving average (EMA) of the student model $\theta$:
\begin{align}
    \theta_t' = \alpha \times \theta_{t-1}' + (1-\alpha) \times \theta_t,
\end{align}
where $t$ denotes the training step and $\alpha$ denotes a smoothing coefficient. 
The output of the teacher network is used to supervise the student network, \ie the output of the refinement block $\mathcal{R}_\mathcal{T}$. We apply a random perturbation $\mathcal{P}$ to the input of the student network, and we denote the output of the teacher network as $T_{\mathcal{T}}$. The random perturbation $\mathcal{P}$ is concurrently applied to the output of the teacher network, \ie $\mathcal{P}\mathcal{T}_\mathcal{T}$. We then enforce the self-consistency loss on the student-teacher network:
\begin{equation}
    \mathcal{L}_\text{mt} = \frac{1}{\mathcal{N}} \sum_{i,j,c}\|\mathcal{R}_\mathcal{T} (i,j,c) - \mathcal{P}T_\mathcal{T} (i,j,c)\|^2.
    \label{eq:LossMT}
\end{equation}
$\mathcal{P}$ is generated from random image rotation, flipping, occlusion, and Gaussion noise. Note that we only apply perturbations that will affect the final output to the teacher network, \ie random rotation and flipping. 
Similar to the self-distillation module, we also concurrently supervise the output of the refinement block $\mathcal{R}_\mathcal{T}$ with the noisy pseudo labels $\Tilde{H}_\mathcal{T}$ via the following loss function:
\begin{equation}
    \begin{split}
        & \mathbf{L}_{\mathcal{R}} = \frac{1}{|\mathcal{C}|} \sum_{c \in \mathcal{C}}\mathcal{L}^c_{\mathcal{R}}, \quad \text{where}~~\mathcal{C} = \{c \mid  \mathcal{L}_{\mathcal{R}}^{c} < l_\text{th}\}, \\
        & \mathcal{L}^c_{\mathcal{R}} = \frac{1}{\mathcal{M}} \sum_{i,j}\|\mathcal{R}_\mathcal{T} (i,j,c) - \Tilde{H}_\mathcal{T} (i,j,c)\|^2.
    \end{split} \label{eq:lossPseudoRefine}
\end{equation}
We use the dynamic threshold $l_\text{th}$ to gradually drop the large loss terms. Similar to the noisy pseudo label on MDAM loss in Eq.~\ref{eq:lossPseudoMDAM}, $\mathcal{L}^c_{\mathcal{R}}$ does not sum over $c$.

It is shown in \cite{Nguyen2020SELFLT} that the teacher network is able to provide more stable learning signal than the pseudo labels since it is a temporal ensemble of networks.
Therefore, we also add a gradually increasing weight term to $\mathcal{L}_\text{mt}$ in the total loss. This means that the outputs of the teacher network are taken to be better pseudo labels to replace the old noisy ones at the later stage of training, and thus preventing overfitting to the noisy pseudo labels. 

\vspace{-3mm}
\paragraph{Remarks:} Note that we place the self-distillation module in the inner loop for coarse update since self-distillation is based on the self-feedback loop with a softer regulatory strength compared to the mean-teacher based on self-consistency. It is beneficial to do the softer self-distillation before the stronger outer loop fine updates by the mean-teacher in the nested loops. The softer regulations from self-distillation prevents the mean-teacher from making drastic replacement of the initial noisy pseudo labels too quickly in the training. Consequently, this allows the network to avoid the \textit{memorization effect} \cite{pmlr-v70-arpit17a, ZhangBHRV17}, and therefore benefit from the noisy pseudo labels at the early stage and then the better pseudo labels at the later stage of training.

\subsection{MixUp Regularizer}
We further adopt the recently proposed MixUp \cite{Zhang2018mixupBE} to further enhance the robustness of our network to the noisy pseudo labels. Specifically, MixUp reduces the negative effect of noisy pseudo labels by combining pseudo labels with the ground truth labels.
%The MixUp \cite{Zhang2018mixupBE} is a recently proposed data augmentation technique, which has been applied to various scenario such as domain adaptation \cite{xu2020adversarial, Wu2020DualMR}, semi-supervised learning \cite{berthelot2019mixmatch, wang2019semi} and learning from noisy labels \cite{ICML2019_UnsupervisedLabelNoise}. 
Given a pair of images $(I_\mathcal{S}, I_\mathcal{T})$ from the source and target domains, and the corresponding ground truth and pseudo label heatmaps $(H_\mathcal{S}, \Tilde{H}_\mathcal{T})$, we perform MixUp to construct virtual training examples by :
\begin{equation}
\begin{split}
    & \lambda \sim \text{Beta}(\alpha, \alpha), \quad \lambda' = \max(\lambda, 1-\lambda), \\
    & I_\mathcal{S}' = \lambda' I_\mathcal{S} + (1-\lambda')I_\mathcal{T}, \\ 
    & H_\mathcal{S}' = \lambda' H_\mathcal{S} + (1-\lambda')\Tilde{H}_\mathcal{T}.
\end{split}
\end{equation}
$\text{Beta}(\alpha, \alpha)$ is the Beta distribution, where we set both hyperparameters to be $\alpha$. $\lambda$ is the parameter to determine the weight of the MixUp from the source and target domains. 
$I'_\mathcal{S}$ and $H'_\mathcal{S}$ are the input image and label heatmap in the source domain after MixUp.
We take the maximum value of $(\lambda, 1-\lambda)$ such that $I_\mathcal{S}'$ is closer to $I_\mathcal{S}$ than to $I_\mathcal{T}$. This is to ensure that the domain label for $I_\mathcal{S}'$ is unchanged after applying MixUp. We also generate virtual example for $I_\mathcal{T}$ by simply changing the max(.,.) to the min(.,.) operator.

%  is a hyperparamter and $H$ represents heatmaps instead of the original joint locations. Note that we take the maximum value of $(\lambda, 1-\lambda)$ as in \cite{berthelot2019mixmatch} such that $I_\mathcal{S}'$ is closer to $I_\mathcal{S}$ than to $I_\mathcal{T}$. This is to ensure that the domain label for $I_\mathcal{S}'$ is unchanged after applying MixUp. We also generate virtual example for $I_\mathcal{T}$ in the same way. We apply MixUp in this work because it exhibits strong robustness to noisy labels. Moreover, by combining pseudo labels with ground truth labels, the negative effect of noisy pseudo labels can be reduced. 

\subsection{Optimization}
The overall objective function to train our network can be expressed as:
\begin{equation}
    \mathcal{L} = \mathcal{L}_{\mathcal{S}} + \lambda_\text{adv} \mathcal{L}_\text{adv} + \mathcal{L}_\text{inner} + \mathcal{L}_\text{outer},
\end{equation}
where
\begin{equation*}
    \begin{split}
        & \mathcal{L}_\text{inner}  = \lambda_\text{sd} \mathcal{L}_\text{sd} + \lambda_{\mathcal{T}}\mathbf{L}_\mathcal{T}, \\
        & \mathcal{L}_\text{outer} =  \lambda_\text{mt} \mathcal{L}_\text{mt} + \lambda_\mathcal{R} \mathbf{L}_\mathcal{R}.
    \end{split}
\end{equation*} 
%   \begin{align}
%      \mathcal{L} = \mathcal{L}_{mdam} + \mathcal{L}_{r},
%  \end{align}
%  where 
%  \begin{align}
%      \mathcal{L}_{mdam} =  \mathcal{L}_\mathcal{S} + \lambda_\mathcal{T} \mathcal{L}_\mathcal{T} + \lambda_\text{adv} \mathcal{L}_\text{adv} + \lambda_\text{sd} \mathcal{L}_\text{sd},
%  \end{align}
%  and 
%  \begin{align}
%      \mathcal{L}_{r} = \lambda_{rp} \mathcal{L}_{rp} + \lambda_c \mathcal{L}_c.
%  \end{align}
$\mathcal{L}_{\mathcal{S}}$ is the fully supervised loss in the source domain (cf. Eq.~\ref{eq: pose_source}) and $\mathcal{L}_\text{adv}$ represents the adversarial loss (cf. Eq.~\ref{eq:lossDomainDisc}). $\mathcal{L}_\text{inner}$ consists of the two loss terms in the inner loop: 1) the self-distillation loss $\mathcal{L}_\text{sd}$ (cf. Eq.~\ref{eq:lossSelfDist}) and 2) noisy pseudo labels on MDAM loss $\mathbf{L}_\mathcal{T}$ (cf. Eq.~\ref{eq:lossPseudoMDAM}). $\mathcal{L}_\text{outer}$ is the two loss terms in the outer loop: 1) mean-teacher loss $\mathcal{L}_\text{mt}$ (cf. Eq.~\ref{eq:LossMT}) and 2) noisy pseudo labels on the refinement block loss $\mathbf{L}_\mathcal{R}$ (cf. Eq.~\ref{eq:lossPseudoRefine}). 
Furthermore, the domain classifier concurrently minimizes $\mathcal{L}_d$ (cf. Eq.~\ref{eq:lossDomainClass}), and the adversarial training is implemented with the gradient reversal layer. 

$\lambda_\text{adv}$, $\lambda_\text{sd}$, $\lambda_\mathcal{T}$, $\lambda_\text{mt}$, $\lambda_\mathcal{R}$ are the weights to balance all losses. As mentioned in the previous section, we gradually increase $\lambda_\text{mt}$ and $\lambda_\text{sd}$ from 0 to their maximum value at the first 10 epochs of training by using a sigmoid-shape function  $e^{-5(1-x)^2}$~\cite{tarvainen2017mean}, where $x \in [0, 1]$. At the same time, we also decrease the $\lambda_\mathcal{T}$ and $\lambda_\mathcal{R}$ at each epoch until to the minimum value.
Note that $\lambda_{\mathcal{T}}$ and $\lambda_\mathcal{R}$ are responsible for balancing the losses, and play no role in removing the noisy pseudo labels in the training. 
The dynamic threshold $l_\text{th}$ in $\mathbf{L}_{\mathcal{T}}$ and $\mathbf{L}_\mathcal{R}$ is responsible for removing noisy pseudo labels. We determine $l_\text{th}$ using Algorithm~\ref{algo:dynThreshold}, where it is dynamically set to the value of the $\alpha_N^\text{th}$ smallest value of $\mathcal{L}^c_\mathcal{T}$ or $\mathcal{L}^c_\mathcal{R}$. $\alpha_N$ is the cut-off index, which we initialize to $K$ and gradually decrease it during training. 
\begin{algorithm}[hb]
% \SetKwData{Left}{left}\SetKwData{This}{this}\SetKwData{Up}{up}
% \SetKwFunction{Concatenate}{Concatenate}\SetKwFunction{Kmeans}{K-means}
\SetKwInOut{Input}{Input}\SetKwInOut{Output}{Output}
\Input{\small{Loss $\mathcal{L}_\text{y}^c$ of $K$ joints $\{ \mathcal{L}_\text{y}^1, \dots, \mathcal{L}_\text{y}^K \}$, where \\ $y = \mathcal{T}$ (cf. Eq.~\ref{eq:lossPseudoMDAM}) or $y = \mathcal{R}$ (cf. Eq.~\ref{eq:lossPseudoRefine})\; \\ Cut-off index $\alpha_N$ }}
\Output{\small{Dynamic threshold $l_\text{th}$}}
\tcp{get indices of $\mathcal{L}_\text{y}^c$ in ascending order }
\small{$\{\text{idx}_1, \dots, \text{idx}_K\} \leftarrow$ sort\_ascending($\{ \mathcal{L}_\text{y}^1, \dots, \mathcal{L}_\text{y}^K \}$)} \;
\tcp{get value of $\mathcal{L}_\text{y}^c$ at $c = \text{idx}_{\alpha_N}$}
\small{$l_\text{th} \leftarrow \mathcal{L}_\text{y}^{\text{idx}_{\alpha_N}}$} \;
\caption{\small{Compute Dynamic Threshold $l_\text{th}$}} \label{algo:dynThreshold}
\end{algorithm}

% $\mathcal{L}_{mdam}$ and $\mathcal{L}_{r}$ are the losses for MDAM and the refinement block, and $\lambda_\mathcal{T}$, $\lambda_c$, $\lambda_\text{sd}$ and $\lambda_\text{adv}$ are the weights to balance all losses. The domain classifier minimize $\mathcal{L}_d$ at the same time, and the adversarial training is implemented with the gradient reversal layer. As mentioned in the previous section, we gradually increase $\lambda_c$ and $\lambda_\text{sd}$ from 0 to their maximum value at the beginning 10 epochs of training by using a sigmoid-shape function  $e^{-5(1-x)^2}$\cite{tarvainen2017mean}, where $x \in [0, 1]$. At the same time, we also decrease the $\lambda_\mathcal{T}$ and $\lambda_{rp}$ at each epoch until to the minimum value. The loss threshold $l_\text{th}$ in $\mathcal{L}_{\mathcal{T}}$ and $\mathcal{L}_{rp}$ are updated dynamically based on the ranking scores of $\mathcal{L}_{\mathcal{T}}$ and $\mathcal{L}_{rp}$. Specifically, we compute the ranking of $\mathcal{L}_{rp}$, expressed as $\mathcal{L}_{rp}^r$ , and $l_\text{th}$ is the $N_k$th element of $\mathcal{L}_{rp}^r$, similarly for $\mathcal{L}_{\mathcal{T}}$. We gradually decrease $N_k$ to filter more pseudo labels during training as it reaches the minimum value.

\section{Experiments}

We use Resnet \cite{he2016deep} as our feature extractor $G$, followed by several deconvolutional layers as the pose estimator $P$. As in \cite{chen2018cascaded}, the intermediate feature maps of the pose estimation module are upsampled and then concatenated. The output is fed into both the domain classifier and the refinement block.  The domain classifier has a fully-convolutional architecture, which consists of six convolutional layers with leaky Relu as the activation function. The refinement block has one bottleneck block followed by one convolutional layer. We first pretrain the pose estimation module on the synthetic dataset for 100 epochs, and then use it to generate pseudo labels for real images. Both synthetic and real data are used to train the whole network for 80 epochs. The learning rate starts at 0.00025 and is decreased using the polynomial decay with power of 0.9 \cite{tsai2018learning}. Our model is optimized with Adam \cite{kingma2014adam} with default parameters in Pytorch. More training details are included in the supplementary materials.

\subsection{Datasets}
We train our network with images and pose annotations $\{I_\mathcal{S}, H_\mathcal{S}\}$ for horse and tiger from the Synthetic Animal dataset \cite{mu2020learning} and real images $I_\mathcal{T}$ from the TigDog dataset\cite{del2015articulated}, and test our model on the test split of the TigDog dataset. We test the generalization capacity of our model on the VisDA2019 dataset, which contains the same animal categories as the TigDog dataset. Moreover, we also test our model on unseen animal categories in the Zebra dataset \cite{zuffi2019three} and the Animal-Pose dataset \cite{cao2019cross}. 

\vspace{-3mm}
\paragraph{Synthetic Animal Dataset:} The dataset contains images for five animal categories, including horse, tiger, sheep hound and elephant, with 10000 images for each animal category.
The texture of animals are randomly genrated from the COCO dataset or from the original CAD models.

\vspace{-3mm}
\paragraph{TigDog Dataset:} The dataset provides keypoint annotations for horse and tiger, where the images are taken from YouTube (for horse) and National Geographic documentaries (for tiger). There are 19 keypoints in the dataset, including eyes, chin, shoulders, legs, hip and neck. We only use the images from this dataset for training and evaluate on 18 keypoints which does not include neck following \cite{mu2020learning}.

\vspace{-3mm}
\paragraph{VisDA2019 Dataset:} The dataset is designed for multi-source domain adaptation and semi-supervised domain adaptation on image classification task. There are in total six domains, including real, sketch, clipart, painting, infograph and quickdraw. \cite{mu2020learning} manually annotates the keypoints for horse and tiger from the sketch, painting and clipart domains. We use this dataset to test the generalization capacity of our approach to unseen domains.

\vspace{-3mm}
\paragraph{Zebra and Animal-Pose Datasets:} The Zebra dataset contains images of Gravy's zebra, which are collected in Kenya with pre-computed bounding boxes. The Animal-Pose dataset contains annotations for five animal categories: dog, cat, horse, sheep and cow. We use these two datasets to test the generalization capacity of our model on unseen animals from unseen domains.

\begin{table*}[ht!]
\centering
\small
\setlength{\tabcolsep}{2.3pt}
\begin{tabular*}{1.0\textwidth}{c|c c c c c c c c|c c c c c c c c}
\hline
& \multicolumn{8}{c|}{Horse Accuracy} & \multicolumn{8}{c}{Tiger Accuracy} \\
 & Eye & Chin & Shoulder & Hip & Elbow & Knee & Hooves & Mean 
& Eye &Chin & Shoulder & Hip & Elbow & Knee & Hooves & Mean \\ \hline
Real  & 79.04  & 89.71 & 71.38 & 91.78 & 82.85 & 80.80 & 72.76 & 78.98 
      & 96.77 & 93.68 & 65.90 & 94.99 & 67.64 & 80.25 & 81.72 & 81.99 \\ \hdashline
Syn & 46.08 & 53.86 & 20.46 & 32.53 & 20.20 & 24.20 & 17.45 & 25.33
      & 23.45 & 27.88 &14.26 & 52.99 & 17.32 & 16.27 & 19.29 & 21.17 \\ 
Cycgan \cite{zhu2017unpaired} & 70.73 & 84.46 & 56.97 & 69.30 & 52.94 & 49.91 & 35.95 & 51.86
            & 71.80 & 62.49 & 29.77 & 61.22 & 36.16 & 37.48 & 40.59 & 46.47 \\
BDL \cite{li2019bidirectional}  & 74.37 & 86.53 & 64.43 &75.65 & 63.04 & 60.18 & 51.96 & 62.33
        & 77.46 & 65.28 & 36.23 & 62.33 & 35.81 & 45.95 & 54.39 & 52.26 \\
Cycada \cite{hoffman2018cycada} & 67.57 & 84.77 & 56.92 & 76.75 & 55.47 & 48.72 &43.08 & 55.57 
        & 75.17 & 69.64 & 35.04 & 65.41 & 38.40 & 42.89 & 48.90 & 51.48 \\
CC-SSL \cite{mu2020learning}  & 84.60 & 90.26 & 69.69 & \textbf{85.89} & 68.58 & 68.73 & 61.33 & 70.77
        & 96.75 & 90.46 & 44.84 & 77.61 & \textbf{55.82} & 42.85 & 64.55 & 64.14 \\ 
% Ours(single) & 88.45 & 93.31 & 70.10 & 82.16 & 71.66 & 77.79 & 71.45 & 77.06 
%             & 99.40 & 92.07 & 49.46 & 71.53 & 53.58 & 62.46 & 72.94 & 69.05 \\ 
Ours & \textbf{91.05} & \textbf{93.37} & \textbf{77.35} & 80.67 & \textbf{73.63} & \textbf{81.83} & \textbf{73.67} & \textbf{79.50}
            & \textbf{97.01} & \textbf{91.18} & \textbf{46.63} & \textbf{78.08} & 50.86 & \textbf{61.54} & \textbf{70.84} & \textbf{67.76} \\ \hline
\end{tabular*}
\caption{PCK@0.05 accuracy for the TigDog dataset. `Real' and `Syn' represent models trained with the labeled real or synthetic dataset, respectively. All other approaches are trained with the labeled synthetic dataset and the unlabeled real dataset. (Best results in bold)}
\label{resuts-tigdog}
\end{table*}

\subsection{Results on the TigDog Dataset}
% We first report our results on the TigDog dataset,  
% and compare with state-of-the-art unsupervised domain adaptation approaches. 
The Percentage of Correct Keypoints (PCK), which reports the percentage of detections that fall within a normalized distance, is used as the evaluation metric following \cite{mu2020learning}. 
We train a unified model on all animal categories instead of training one model for each animal category as in \cite{mu2020learning}. We believe that this is more practical in real setting. 
The PCK@0.05 accuracy of our approach and existing unsupervised domain adaptation approaches, which are taken from \cite{mu2020learning}, are shown in Tab.~\ref{resuts-tigdog}. The `Real' represents model trained with the real animal pose data and `Syn' represents model trained only with synthetic data.
As can be seen from Tab.~\ref{resuts-tigdog}, our model outperforms existing unsupervised domain techniques by a large margin. For horse category, our approach improves the state-of-the-art CC-SSL by 12.34\% , and even outperform the model trained with real data. For tiger category, we also achieve the best performance among other UDA techniques with an improvement of 5.64\% comparing to CC-SSL. We did not outperform the supervised model for tiger. The reason is that tigers generally live in forests, where occlusion by surrounding floras happens frequently. However, this kind of occlusion do not occur in the synthetic data, and thus making it very challenging for our model to adapt to the severe occlusion scenario. This also explains why all UDA methods in Tab.~\ref{resuts-tigdog} show better performance for horse, which lives in the grasslands with lesser occlusions.

\subsection{Generalization to Unseen Domains}
We test the generalization capacity of our model by directly applying it to the unseen domains in the VisDA2019 dataset. 
The PCK@0.05 accuracy of our approach for horse and tiger under sketch, painting and clipart domains are shown in Tab.~\ref{results-visda2019}. Following \cite{mu2020learning}, we evaluate our model under two settings: 1) The Visible Kpts Accuracy represents accuracy for only visible joints, and 2) the Full Keypoints Accuracy represents accuracy for all joints including self-occluded joints.  
Both CC-SSL and our approach outperform the model trained on real images, which demonstrates the importance of learning from other domains. Furthermore, our approach also outperforms CC-SSL by a large margin, especially for horse under the painting domain (80.05 \vs 73.71, 78.42 \vs 71.56) and for tiger under all domains. We also show some qualitative results for horse and tiger in each domain in the first row of Fig.~\ref{fig:qualitative}.

\begin{table*}[h!]
\centering
\small
\setlength{\tabcolsep}{4.8pt}
\begin{tabular*}{0.99\textwidth}{c|c c c c c c|c c c c c c}
\hline
    & \multicolumn{6}{c}{Horse} & \multicolumn{6}{c}{Tiger} \\
    \cline{2-13}
     &  \multicolumn{3}{c|}{Visible Kpts Accuracy} & \multicolumn{3}{c|}{Full Kpts Accuracy}
     &  \multicolumn{3}{c|}{Visible Kpts Accuracy} & \multicolumn{3}{c}{Full Kpts Accuracy} \\
     & Sketch & Painting & \multicolumn{1}{c|}{Clipart} & Sketch & Painting & Clipart
     & Sketch & Painting & \multicolumn{1}{c|}{Clipart} & Sketch & Painting & Clipart \\
     \hline
     Real & 65.37 & 64.45 & \multicolumn{1}{c|}{64.43} & 61.28 & 58.19 & 60.49 
            & 48.10 & 61.48 & \multicolumn{1}{c|}{53.36} & 46.23 & 53.14 & 50.92 \\
            
    CC-SSL \cite{mu2020learning} & 72.29 & 73.71 & \multicolumn{1}{c|}{73.47} & 70.31 & 71.56 & 72.24 
            & 53.34 & 55.78 & \multicolumn{1}{c|}{59.34} & 52.64 & 48.42 & 54.66  \\
            
    Ours  & \textbf{76.65}  & \textbf{80.05} & \multicolumn{1}{c|}{\textbf{75.45}} & \textbf{73.74} &  \textbf{78.42} & \textbf{73.61} 
            & \textbf{60.85} & \textbf{61.54} & \multicolumn{1}{c|}{\textbf{65.12}} & \textbf{59.58} & \textbf{56.09} & \textbf{60.66} \\ \hline
\end{tabular*}
\caption{PCK@0.05 accuracy for the VisDA2019 dataset. (Best results in bold) }
\label{results-visda2019}
\end{table*}

\subsection{Generalization to Unseen Animals from Unseen Domains}

We further test the generalization capacity of our model in a more challenging scenario, where our model is directly applied to unseen animal categories from unseen domains. Note that our model is trained only with horse and tiger categories, and we test on both the Zebra dataset and the Animal-Pose dataset. 

\begin{table*}[ht!]
\centering
\small
\setlength{\tabcolsep}{12.8pt}
\begin{tabular*}{0.93\textwidth}{c|c c c c c c c c}
\hline
     & Eye & Chin & Shoulder & Hip & Elbow & Knee & Hooves & Mean \\ \hline
  Zebra3D*\cite{zuffi2019three}   & - & - & - & - & - & - & - & 59.5 \\ \hdashline
  CC-SSL \cite{mu2020learning}  & 60.06 & 82.29 & \textbf{30.30} & 0 & 32.45 & 65.13 & 61.97 & 50.07 \\
  Ours   & \textbf{65.33} & \textbf{87.50} & 23.74 & 0 & \textbf{45.32} & \textbf{76.02} & \textbf{69.77} & \textbf{57.23} \\ \hline
\end{tabular*}
\caption{PCK@0.05 accuracy for the Zebra dataset. * denotes approaches trained with the zebra category. (Best results in bold) }
\label{results-zebra}
\vspace{2mm}
\end{table*}

The Zebra dataset contains images of Gravy's zebra collected in Kenya, and 28 keypoints are provided with each image . We only test on the 18 keypoints that are described in the TigDog dataset. The PCK@0.05 accuracy of our proposed approach is shown in Tab.~\ref{results-zebra}. Zebra3D represents the approach used in \cite{zuffi2019three} for 3D zebra pose estimation. This model is trained on a synthetic zebra dataset, which is generated from zebra models with appearance taking from real zebra images. We compare with their results without the post optimization process. The results of CC-SSL are obtained by running their public available checkpoint. As they train one model for each animal category, we use the one that gives better accuracy on this dataset. We can see that our approach outperforms CC-SSL with an improvement of 14.3\%. Our approach also achieves comparable results to Zebra3D although our model has not been trained on the zebra category. Note that the accuracy of our approach and CC-SSL for joint hip is zero because the joint locations for hip are defined differently for the Synthetic Animal dataset (which is used to train our model) and the Zebra dataset. This is another reason why our approach and CC-SSL are not as good as Zebra3D. 

\begin{figure*}[h!]
    \centering
    
    \includegraphics[width=0.15\textwidth]{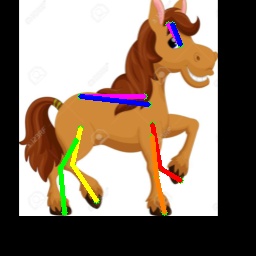}
    \hfill
   \includegraphics[width=0.15\textwidth]{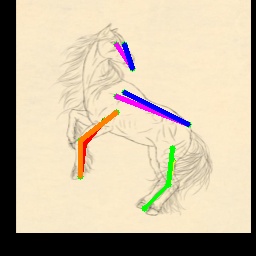}
    \hfill
    \includegraphics[width=0.15\textwidth]{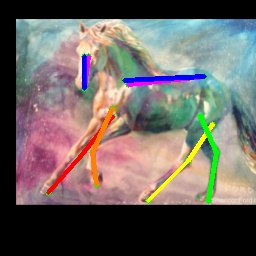}
    \hfill
    \includegraphics[width=0.15\textwidth]{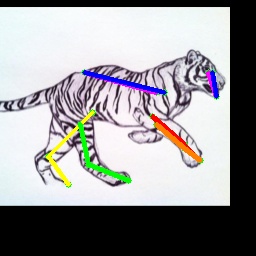}
    \hfill
    \includegraphics[width=0.15\textwidth]{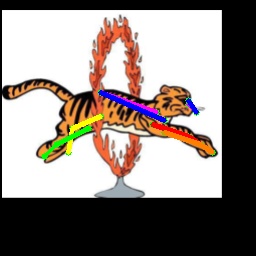}
    \hfill
    \includegraphics[width=0.15\textwidth]{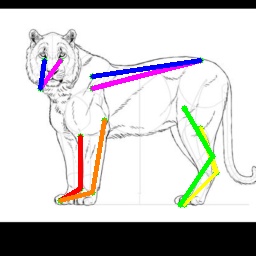} \\ 
        
    \includegraphics[width=0.15\textwidth]{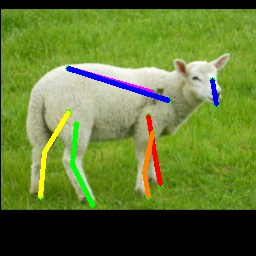}
         \hfill
    \includegraphics[width=0.15\textwidth]{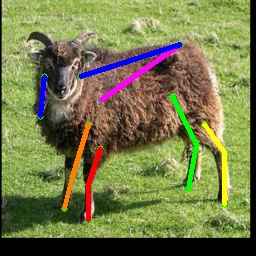}
        \hfill
    \includegraphics[width=0.15\textwidth]{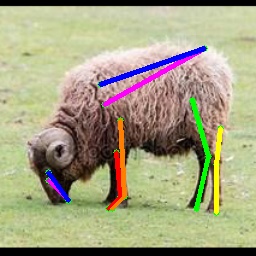}
        \hfill
  \includegraphics[width=0.15\textwidth]{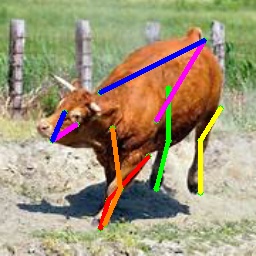}
    \hfill
    \includegraphics[width=0.15\textwidth]{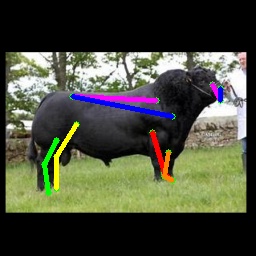}
    \hfill
    \includegraphics[width=0.15\textwidth]{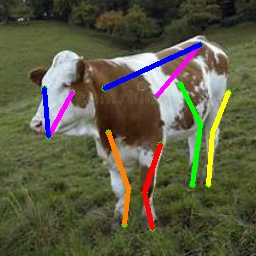} \\ 

    \includegraphics[width=0.15\textwidth]{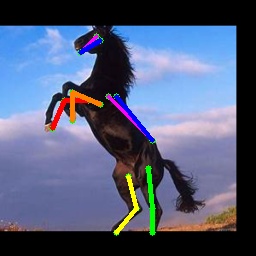}
    \hfill
    \includegraphics[width=0.15\textwidth]{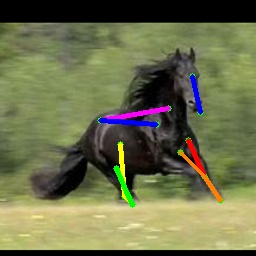}
    \hfill
    \includegraphics[width=0.15\textwidth]{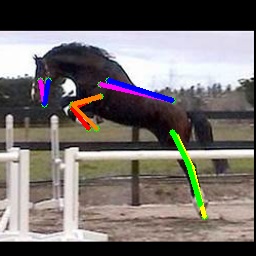}
        \hfill
    \includegraphics[width=0.15\textwidth]{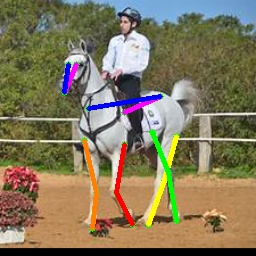}
            \hfill
    \includegraphics[width=0.15\textwidth]{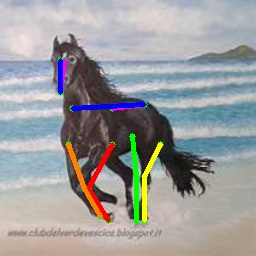} 
    \hfill
    \includegraphics[width=0.15\textwidth]{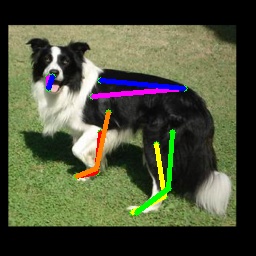}
    \\    

    \includegraphics[width=0.15\textwidth]{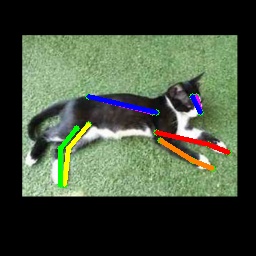}
    \hfill
    \includegraphics[width=0.15\textwidth]{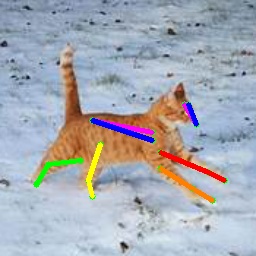}
    \hfill
    \includegraphics[width=0.15\textwidth]{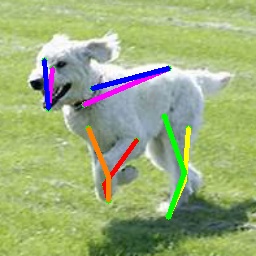}
        \hfill
    \includegraphics[width=0.15\textwidth]{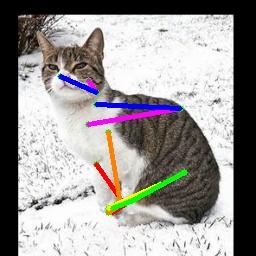}
    \hfill
    \includegraphics[width=0.15\textwidth]{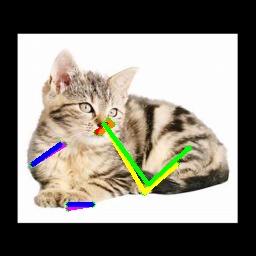}
    \hfill
    \includegraphics[width=0.15\textwidth]{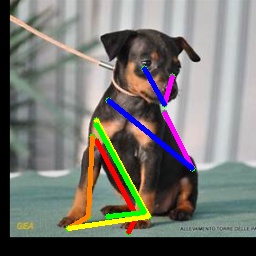} \\
    \caption{Qualitative results for the VisDA2019 dataset (the first row) and the Animal-Pose dataset (the last three rows). 
    % More qualitative results are shown in the supplementary materials.}
    }
    \vspace{-2mm}
    \label{fig:qualitative}
\end{figure*}

We also test on the 1000 images from the Animal-Pose dataset, with 200 images for each animal category. All animal categories in this dataset are unseen except for horse.
We show our results in Tab.~\ref{results-animalpose}, where the results for CC-SSL are from the checkpoint that gives better average accuracy.
% We do not directly compare with \cite{cao2019cross} here because they did not release the code and the validation split. 
We can see that our approach can generalize well to unseen animal categories such as sheep and cow, with an accuracy close to horse. The performance of our model for dog and cat is not as good as that for sheep and cow. We attribute this to two reasons: 1) The shape and size of dogs and cats are very differ from horses (or tigers), especially for cats with much smaller size. 2) Dog and cat are always in a sit or prone pose, which is not the case for horse or tiger living in the wild environment. We show some failed examples in Fig.~\ref{fig:qualitative} for illustration (the last three examples in the last row). We also show qualitative results for each animal category in Fig.~\ref{fig:qualitative}. We can see that our model successfully estimates some challenging poses, such as the jumping horse, the lying down cat and the running dog. 
\begin{table}[h!]
\centering
\small
\setlength{\tabcolsep}{4.2pt}
\begin{tabular*}{0.48\textwidth}{c|c c c c c c}
\hline
     & Horse & Dog & Cat & Sheep & Cow & Mean \\ \hline
  CC-SSL \cite{mu2020learning}  & 65.35 & 30.27 & 15.05 & 52.39 & 63.71 & 47.60\\
  Ours   & \textbf{72.84} & \textbf{42.48} & \textbf{27.65} & \textbf{59.51} & \textbf{71.31} & \textbf{56.77} \\ \hline
\end{tabular*}
\caption{PCK@0.05 accuracy for the Animal-Pose dataset. All animal category are unseen except for horse.}
\vspace{-3mm}
\label{results-animalpose}
\end{table}

\subsection{Ablation Study}
We conduct ablation study on the TigDog dataset and the results are shown in Tab.~\ref{ablation}. We use the multi-scale domain adaptation module as our backbone architecture and train it with only the pseudo labels (mdam+pl) or the supervision from the teacher network (mdam+mt). We also compare with CC-SSL\cite{mu2020learning} where the authors train the model and update the pseudo label in an iterative way.  We can see that our backbone MDAM outperforms CC-SSL because we explicitly enforce the network to learn domain invariant features by applying a domain classifier. The MDAM trained with the teacher network is not as good as the one trained with pseudo labels, suggesting that the teacher network alone can not provide enough supervision. The performance is improved by adding the outer fine-update loop (mdam+mt+outlp), where we gradually update the pseudo labels with the teacher network. This demonstrates the importance of our progressive updating strategy, which helps the network learn from pseudo labels at the early stage and then from the more accurate teacher network. Moreover, the performance is further improved by adding the inner coarse-update loop (mdam+mt+outerlp+inlp). This shows the efficiency of updating the pseudo labels in a coarse-to-fine manner. Finally, our model is further enhanced with the MixUp regularizer (full model). 
\begin{table}[ht!]
\centering
\small
\setlength{\tabcolsep}{12pt}
\begin{tabular*}{0.48\textwidth}{c|c c c}
\hline
     & Horse & Tiger & Mean \\ \hline
  CC-SSL \cite{mu2020learning}  & 70.77 & 64.14 & 67.52\\
  mdam + pl   & 74.42 & 64.90 & 69.69 \\
  mdam + mt  & 74.74 & 62.62 & 68.70\\
  mdam + mt + outlp   & 78.38 & 67.15 & 72.70\\
  mdam + mt + outlp + inlp  & 78.53 & 68.01 & 73.25\\
  full model   & 79.50 & 67.76 & 73.66 \\ \hline
\end{tabular*}
\caption{Ablation study for each component of our network.}
\vspace{-5mm}
\label{ablation}
\end{table}

\section{Conclusion}
We propose an approach for unsupervised domain adaptation on animal pose estimation. A multi-scale domain adaptation module is designed to transfer knowledge from the synthetic source domain to the real target domain. 
In addition, a coarse-to-fine pseudo label updating strategy is further introduced to gradually replace noisy pseudo labels with more accurate ones during training. As a result, we enable our network to benefit from the noisy pseudo labels at the early stage, and the updated labels at the later stage without suffering from the ``memorization effect". Extensive experiments on several benchmark datasets show the effectiveness of our approach. 

\section*{Acknowledgement}
This work is supported by the Tier 2 grant MOE-T2EP20120-0011 from the Singapore Ministry of Education.
{\small
\bibliographystyle{ieee_fullname}
\bibliography{egbib}
}
% \end{document}

% \vfill\eject
\pagebreak

% \begin{document}

%%%%%%%%% TITLE
	\twocolumn[{%
		\renewcommand\twocolumn[1][]{#1}%
		\vskip .5in
		\begin{center}
			\textbf{\Large From Synthetic to Real: Unsupervised Domain Adaptation}\\
			\vspace*{4pt}
			\textbf{\Large for Animal Pose Estimation}\\
			\vspace*{6pt}
			\textbf{\Large (Supplementary Material)}\\
			% additional two empty lines at the end of the title
			\vspace*{10pt}
			{\large
				Chen Li \quad \quad Gim Hee Lee\\
			}
			% additional small space at the end of the author name
			\vskip .5em
			{\large Department of Computer Science, National University of Singapore\\}
			{\tt\small \{lic, gimhee.lee\}@comp.nus.edu.sg}
			\vspace*{10pt}
		\end{center}
	}]

\setcounter{equation}{0}
\setcounter{figure}{0}
\setcounter{table}{0}
\setcounter{section}{0}
\setcounter{page}{1}

%%%%%%%%% BODY TEXT
\paragraph{Implementation details.}
We train our network in two stages. We first use the synthetic data to pretrain the pose estimation module, which is used to generate pseudo labels for the second stage. The model is trained for 100 epochs with a learning rate of 0.00025, which decays twice at 60 and 90 epochs respectively. The pseudo labels and the corresponding confidence scores are generated in the same way as in \cite{mu2020learning}. We then use both the synthetic data and pseudo labels to train our full model. Note that we filter out noisy samples with low confidence instead of using all pseudo labels. $\lambda_{sd}$ and $\lambda_{mt}$ increase gradually from 0 to 90 in the first 10 epochs of training. On the other hand, $\lambda_\mathcal{T}$ and $\lambda_\mathcal{R}$ decrease gradually from 15 to 8 in the first 15 epochs. We also remove samples with large loss values at the same time. The cut-off index $\alpha_N$ in Algorithm 1 in our main paper is set to 18 (\ie the number of joints) at the beginning, and then decreases by one after each epoch until it reaches the minimum value 9. We apply random flip,  rotation  (-45$^\circ$ - 45$^\circ$), scale (0.6 - 1.4), noise (-0.2 - 0.2), and random occlusion to the input image as data augmentation.

\begin{figure*}[h!]
\centering
\includegraphics[width=0.3\textwidth]{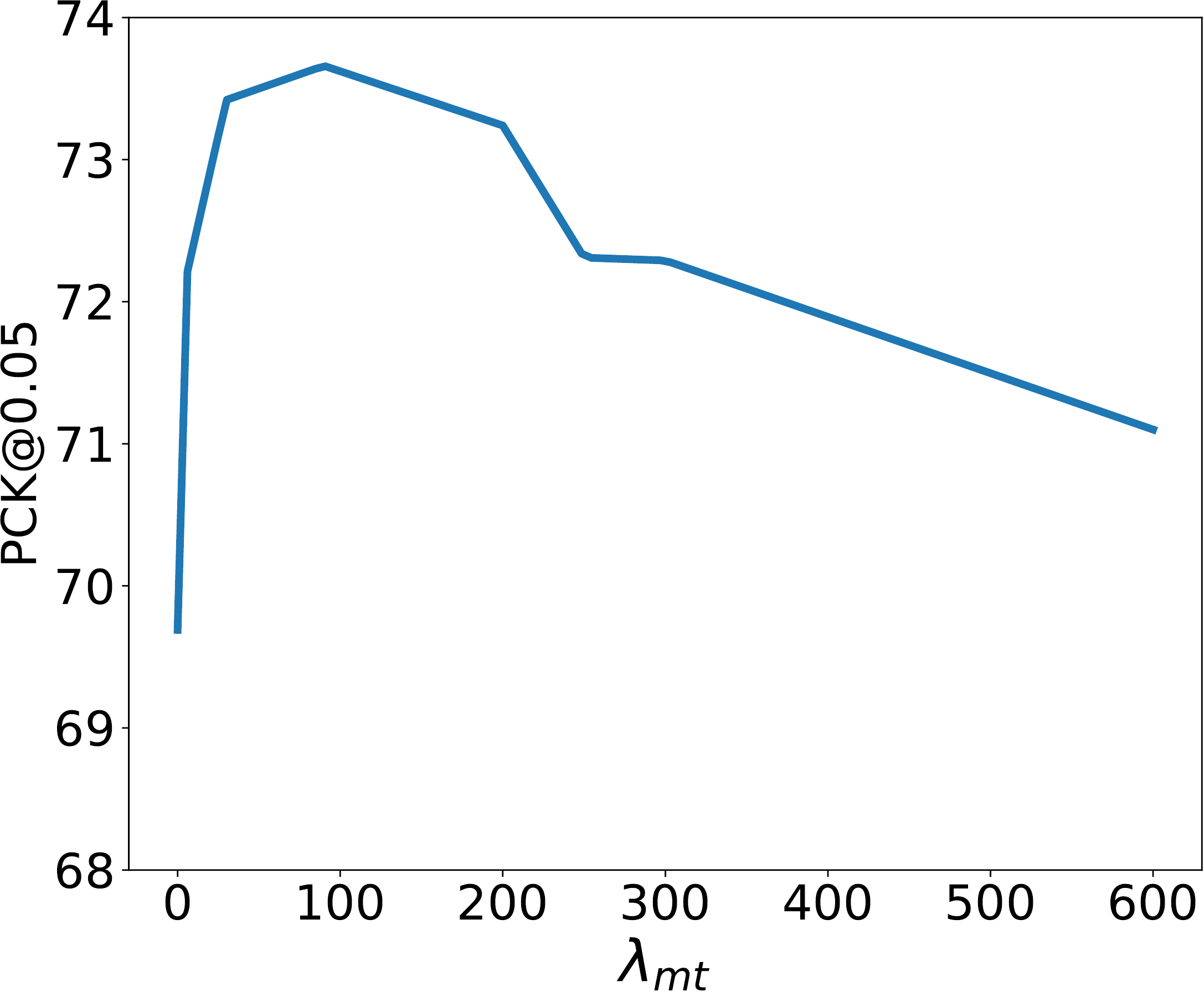} \quad
\includegraphics[width=0.3\textwidth]{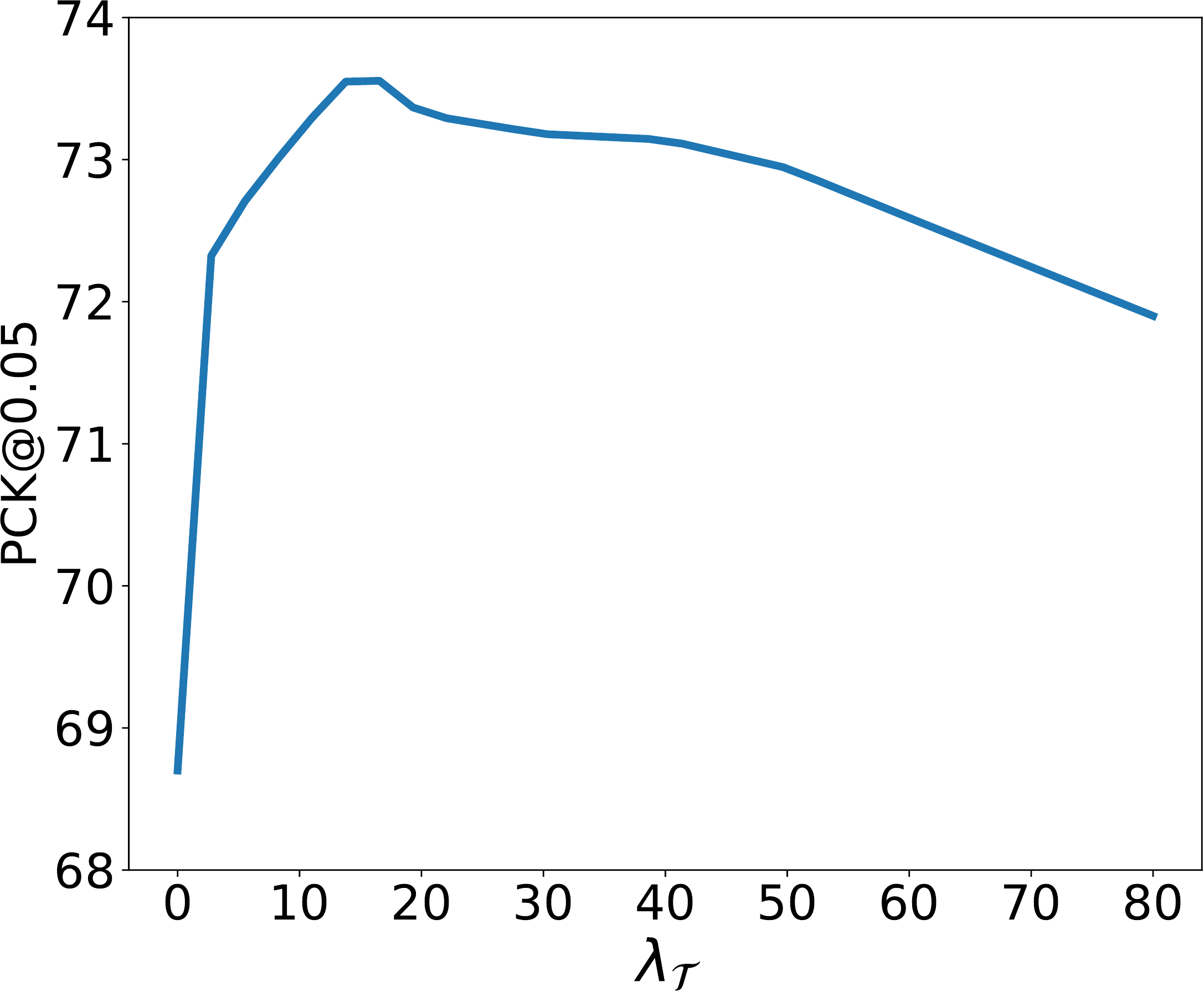} \quad
\includegraphics[width=0.3\textwidth]{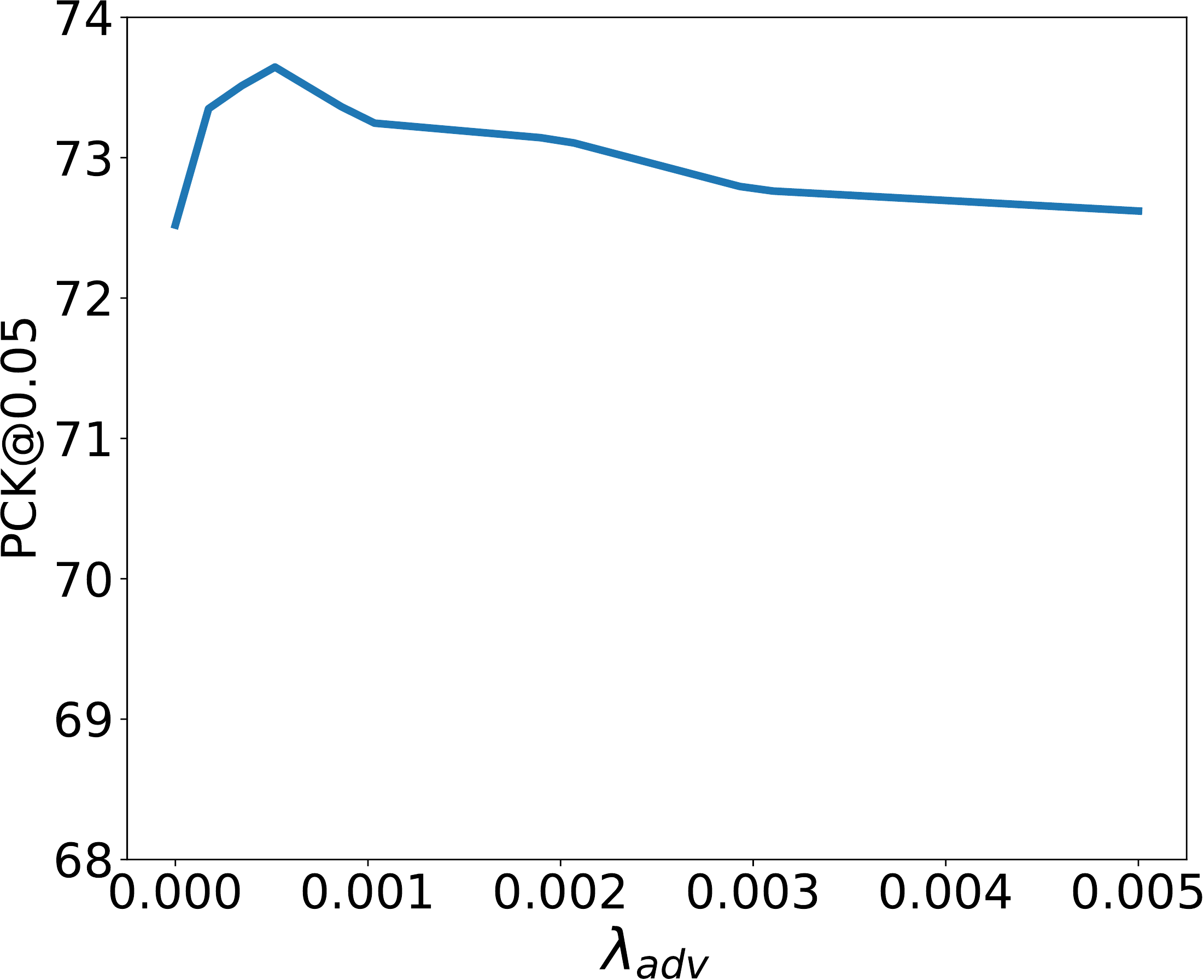} 
\caption{Sensitivity analysis on weight terms $\lambda_\text{mt}$ (left), $\lambda_\mathcal{T}$ (middle) and $\lambda_{adv}$(right).}
\label{hyperparameter-analysis}
\end{figure*}

\paragraph{Analysis on hyperparameters.}
We conduct sensitivity analysis on the hyperparameters we used in the overall objective function (\cf Eq.~12). 
Note that We set $\lambda_\text{sd}=\lambda_\text{mt}$ and $\lambda_\mathcal{T}=\lambda_\mathcal{R}$ since they are counterparts in the inner and outer loops, respectively. Hence, we 
only need to tune three weight terms instead of five. We perform analysis over a range of hyperparameters: $\lambda_\text{sd}=\lambda_\text{mt}=[0,600]$,  $\lambda_\mathcal{T}=\lambda_\mathcal{R}=[0,80]$ and $\lambda_{adv}=[0,0.005]$ on the TigDog dataset. As shown in Fig.~\ref{hyperparameter-analysis}, our model shows: 1) performance of initial increase and then flattened at $\lambda_\text{sd}=\lambda_\text{mt}=[50, 200]$ before finally decreasing performance. 2) Initial increase and then flattened at $\lambda_\mathcal{T}=\lambda_\mathcal{R}=[10, 40]$ before finally decreasing performance. 3) Initial increase and then flattened at $\lambda_{adv}=[0.0001, 0.002]$ before finally decreasing performance. We select the best performing $\lambda_\text{sd}=\lambda_\text{mt}=90$, $\lambda_\mathcal{T}=\lambda_\mathcal{R}=15$ and $\lambda_{adv}=0.0005$ for all our experiments.

\begin{figure}[ht!]
\centering
\includegraphics[width=0.4\textwidth]{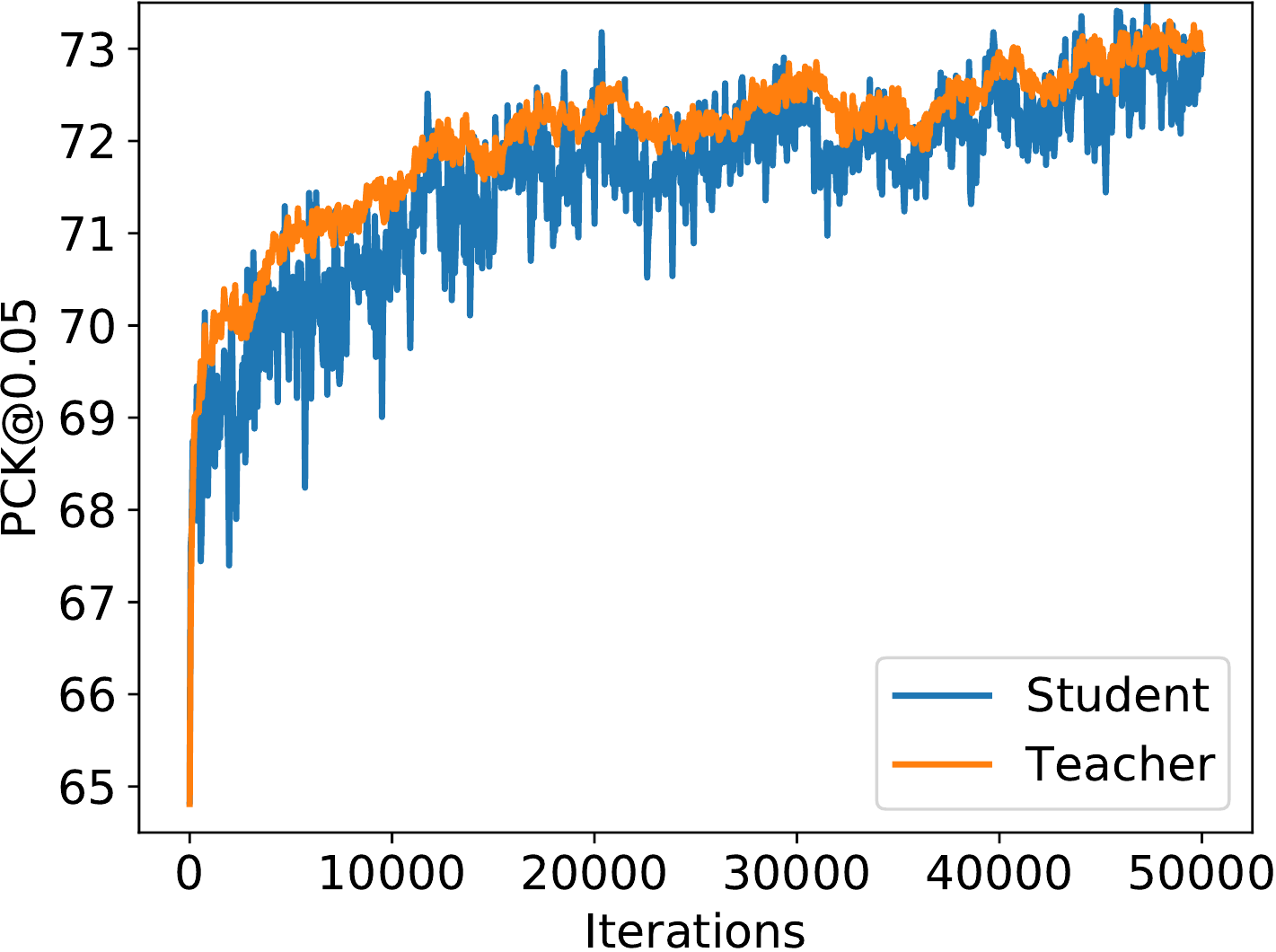}
\caption{The accuracy of the student and teacher network over 50,000 iterations of training.}
\label{pseudolabel-analysis}
\end{figure}

\paragraph{The role of the teacher network.} The role of the teacher network is to provide more stable training signal for the student network. To verify this, we show the accuracy of the student and teacher network over 50,000 iterations of training in Fig.~\ref{pseudolabel-analysis}. We can see that the accuracy of the teacher network improves fast in the first 12,000 iterations and then gradually slows down. This means that the quality of the pseudo labels generated by the teacher network is gradually improved as training proceeds. Moreover, the teacher network is more stable compared to the student network, especially at the beginning of the training. We can also observe from Fig.~\ref{pseudolabel-analysis} that the student network gradually becomes stable as the more and more pseudo labels are updated by the teacher network.  

\paragraph{Baseline using just synthetic horse or tiger.}We train one model for all animal categories instead of training one model for each animal category as in \cite{mu2020learning} in our main paper. To verify that our improved results does not come from the combination of data, we also train one model for each animal category and compare with the model trained with all animal categories. The comparative results are 77.06 (trained only on horse) vs 79.50 (trained on both horse \& tiger) for horse and 69.05 (trained only on tiger) vs 67.76
(trained on both horse \& tiger) for tiger.
This ablation study shows that our model gives 
comparable average accuracy
when trained on all categories or a single category. 

\paragraph{Architecture for the domain classifier.} The domain classifier has a fully-convolutional architecture. Specifically, there are six convolutional layers, with kernel size of $4 \times 4$ and stride of 2 for the first five layers, and with kernel size of $2 \times 2$ and stride of 1 for the last layer. The number of channels is $\{64, 128, 256, 512, 1024, 1\}$ and each convolutional layer is followed by a leaky Relu \cite{maas2013rectifier} except for the last layer. 

\paragraph{Architecture for the pose estimator.} We utilize the Resnet50 to extract the feature maps, which are fed into a pose estimator to generate the output heatmaps. The pose estimator consists of one convolutional layer, followed by three deconvolutional layers, and another convolutional layer. The first $1 \times 1$ convolutional layer has channel size of 256. Each deconvolutional layer has $256$ filters, with kernel size of $4 \times 4$ and stride of 2. An $1 \times 1$ convolutional layer with channel size of $K$ is added at last to generate heatmaps for all $K$ joints. Batch normalization \cite{pmlr-v37-ioffe15} and Relu activation \cite{NIPS2012_c399862d} are applied after each layer except for the last layer.

\paragraph{Architecture for the refinement block.} The input of the refinement block is the concatenation of intermediate feature maps of the pose estimator. The feature maps are fed into a bottleneck block followed by a $3 \times 3$ convolutional layer to generate the final outputs. 

\begin{figure*}[ht!]
    \centering
    \includegraphics[width=0.15\textwidth]{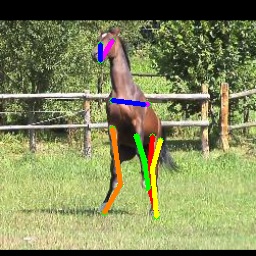}
    \hfill
    \includegraphics[width=0.15\textwidth]{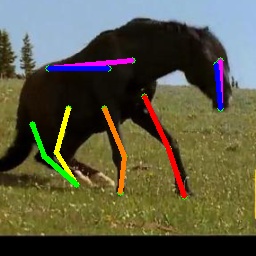}
    \hfill
    \includegraphics[width=0.15\textwidth]{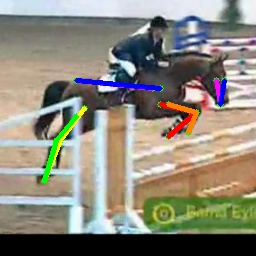}
    \hfill
      \includegraphics[width=0.15\textwidth]{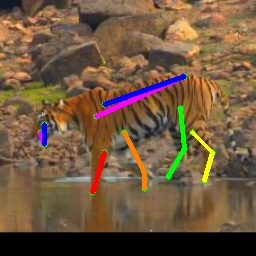}
    \hfill
        \includegraphics[width=0.15\textwidth]{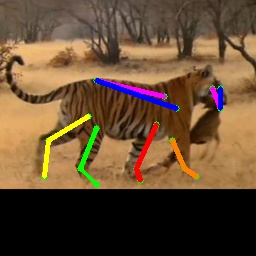}
    \hfill
    \includegraphics[width=0.15\textwidth]{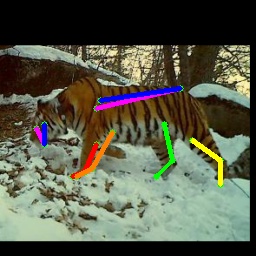}
    \\ 
    % \includegraphics[width=0.15\textwidth]{latex/images/2945.jpg}
    % \hfill
    % \includegraphics[width=0.15\textwidth]{latex/images/2969.jpg}
    % \hfill
    % \includegraphics[width=0.15\textwidth]{latex/images/3065.jpg}
    % \hfill

    \includegraphics[width=0.15\textwidth]{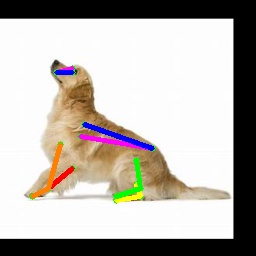}
    \hfill
    \includegraphics[width=0.15\textwidth]{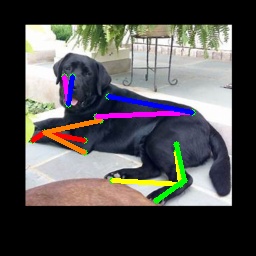}
    \hfill
    \includegraphics[width=0.15\textwidth]{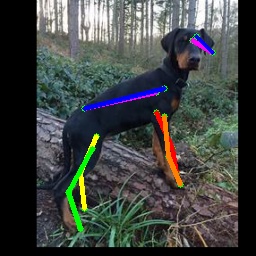}
        \hfill    
    \includegraphics[width=0.15\textwidth]{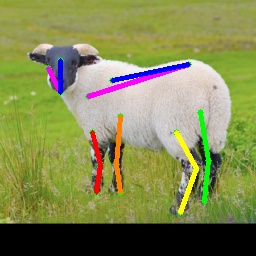}
    \hfill
    \includegraphics[width=0.15\textwidth]{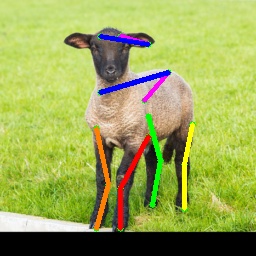}
    \hfill
        \includegraphics[width=0.15\textwidth]{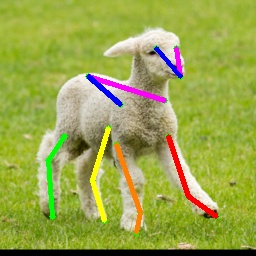}
    \\

    \includegraphics[width=0.15\textwidth]{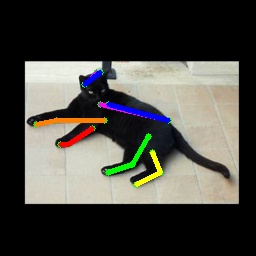}
    \hfill
    \includegraphics[width=0.15\textwidth]{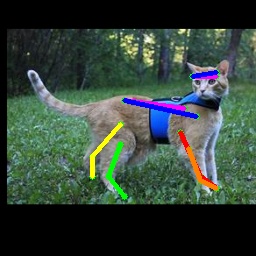}
    \hfill
    \includegraphics[width=0.15\textwidth]{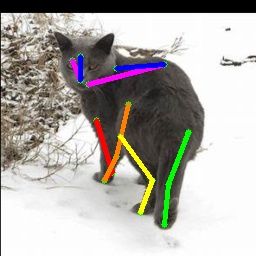}
    \hfill
    \includegraphics[width=0.15\textwidth]{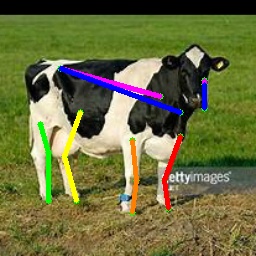}
        \hfill
    \includegraphics[width=0.15\textwidth]{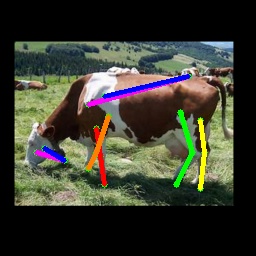}
    \hfill    
    \includegraphics[width=0.15\textwidth]{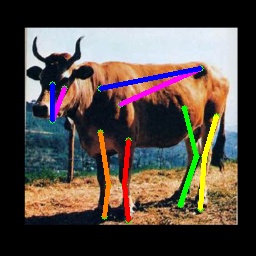}
    \\

    % \includegraphics[width=0.15\textwidth]{latex/images/ho148.jpeg}
    % \hfill
    % \includegraphics[width=0.15\textwidth]{latex/images/ho151.jpeg}
    % \hfill
    % \includegraphics[width=0.15\textwidth]{latex/images/ho63.jpeg}
    % \\
    
    % \includegraphics[width=0.15\textwidth]{latex/images/ho169.jpeg}
    %     \hfill
    % \includegraphics[width=0.15\textwidth]{latex/images/ho58.jpeg}
    % \hfill
    % \includegraphics[width=0.15\textwidth]{latex/images/ho82.jpeg}
    % \hfill
    % \includegraphics[width=0.15\textwidth]{latex/images/ho62.jpeg}
    % \hfill

    % \includegraphics[width=0.15\textwidth]{latex/images/co118.jpeg}
    % \\

    % \includegraphics[width=0.15\textwidth]{latex/images/co168.jpeg}
    %     \hfill
    % \includegraphics[width=0.15\textwidth]{latex/images/co176.jpeg}
    %         \hfill
    % \includegraphics[width=0.15\textwidth]{latex/images/co190.jpeg}
    % \hfill
    % \includegraphics[width=0.15\textwidth]{latex/images/co44.jpeg}
    % \hfill
    % \includegraphics[width=0.15\textwidth]{latex/images/co74.jpeg}
    % \hfill
    
    % \includegraphics[width=0.15\textwidth]{latex/images/clipart_148_000167.jpg} 
    % \includegraphics[width=0.15\textwidth]{latex/images/clipart_148_000027.jpg}
        % \hfill
    % \includegraphics[width=0.15\textwidth]{latex/images/clipart_148_000008.jpg}
    % \hfill        
            
    \includegraphics[width=0.15\textwidth]{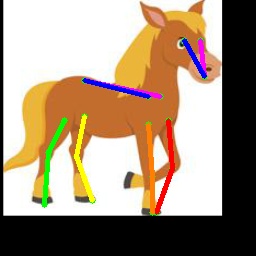} 
    \hfill
    \includegraphics[width=0.15\textwidth]{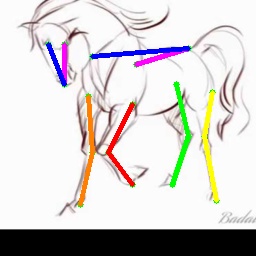}
    \hfill
    \includegraphics[width=0.15\textwidth]{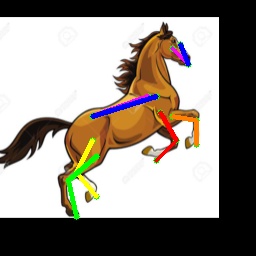}
    \hfill
    \includegraphics[width=0.15\textwidth]{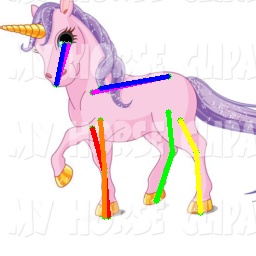}
            \hfill
    \includegraphics[width=0.15\textwidth]{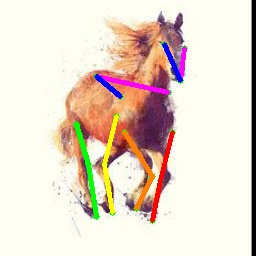} 
        \hfill
    \includegraphics[width=0.15\textwidth]{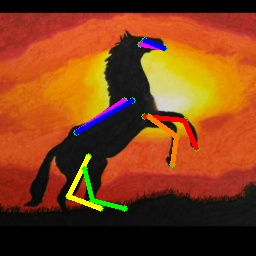} 
    \\
    
    \includegraphics[width=0.15\textwidth]{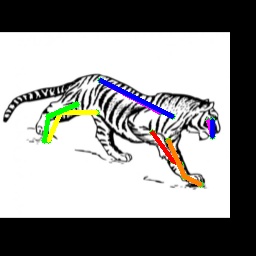}
    \hfill
    \includegraphics[width=0.15\textwidth]{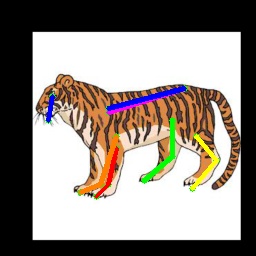}
    \hfill
    \includegraphics[width=0.15\textwidth]{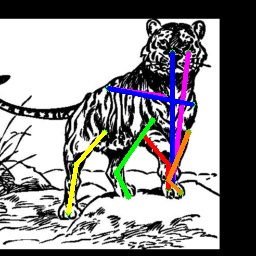}
    \hfill
    \includegraphics[width=0.15\textwidth]{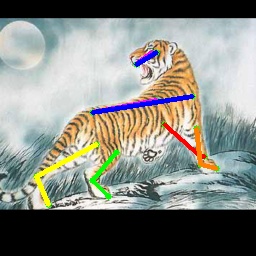}
    \hfill
    \includegraphics[width=0.15\textwidth]{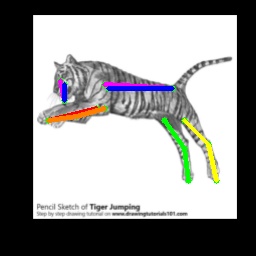}
    \hfill
    \includegraphics[width=0.15\textwidth]{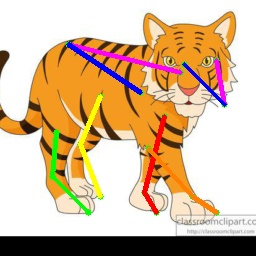}
    \\
    
    \includegraphics[width=0.15\textwidth]{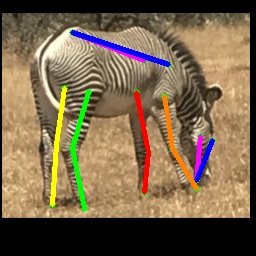}
    \hfill
    \includegraphics[width=0.15\textwidth]{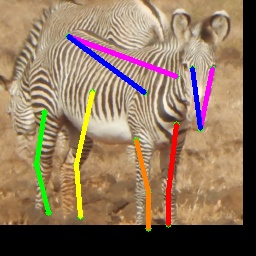}
    \hfill
    \includegraphics[width=0.15\textwidth]{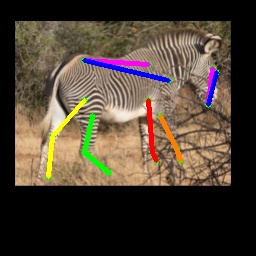}
    \hfill
    \includegraphics[width=0.15\textwidth]{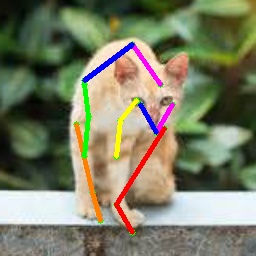}
    \hfill
    \includegraphics[width=0.15\textwidth]{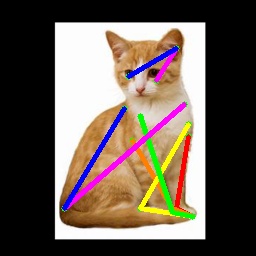}
    \hfill    
    \includegraphics[width=0.15\textwidth]{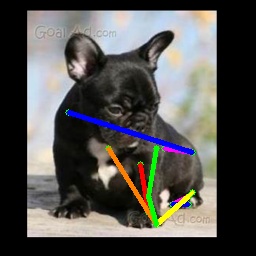}
    \\

    \caption{Qualitative results for the TigDog,Animal-Pose dataset,VisDA2019 dataset and Zebra datasets. The last three examples are the failure cases.}
    \label{fig:qualitative-2}
\end{figure*}

\paragraph{Qualitative results.} We show more qualitative results for the TigDog, Animal-Pose, VisDA2019 and Zebra datasets in Fig.~\ref{fig:qualitative-2}. Our model trained on the synthetic dataset is able to generalize well to the in-the-wild images in the TigDog dataset (the first row). Moreover, our model can also generalize well to the unseen domains in the VisDA2019 dataset (the fourth to the fifth row) and the unseen animal categories in the Animal-Pose dataset (the second to the third row) and the Zebra dataset (the first three images in the last row). We also show failure cases in the last three images, where there are severe self-occlusions. More qualitative results for both seen and unseen categories are included in our video. 

% 	{\small
% 		\bibliographystyleNew{ieee_fullname}
% 		\bibliographyNew{egbib}
% 	}
\end{document}